%% file: all_arxiv.tex
\documentclass[10pt,twocolumn,letterpaper]{article}

\usepackage{cvpr}              %

\usepackage{graphicx}
\usepackage{amsmath}
\usepackage{amssymb}
\usepackage{booktabs}
\usepackage{multirow}
\usepackage{array}
\usepackage{subfloat} 
\usepackage[accsupp]{axessibility}

\usepackage{xcolor}

\usepackage[pagebackref,breaklinks,colorlinks]{hyperref}

\usepackage[capitalize]{cleveref}
\crefname{section}{Sec.}{Secs.}
\Crefname{section}{Section}{Sections}
\Crefname{table}{Table}{Tables}
\crefname{table}{Tab.}{Tabs.}

\def\cvprPaperID{10128} %
\def\confName{CVPR}
\def\confYear{2023}

\begin{document}

\title{NeFII: Inverse Rendering for Reflectance Decomposition with \\ Near-Field Indirect Illumination}

\author{Haoqian Wu$^{1}$, Zhipeng Hu$^{1,2}$, Lincheng Li$^{1}$\footnotemark[1], Yongqiang Zhang$^{1}$, Changjie Fan$^{1}$, Xin Yu$^{3}$\\
$^{1}$ NetEase Fuxi AI Lab $^{2}$ Zhejiang University $^{3}$ The University of Queensland\\
{\tt\small \{wuhaoqian, zphu, lilincheng, zhangyongqiang02, fanchangjie\}@corp.netease.com} \\
{\tt\small xin.yu@uq.edu.au }
}

\maketitle

\renewcommand{\thefootnote}{\fnsymbol{footnote}}
\footnotetext[1]{Corresponding author.}

\begin{abstract}

Inverse rendering methods aim to estimate geometry, materials and illumination from multi-view RGB images.
In order to achieve better decomposition, recent approaches attempt to model indirect illuminations reflected from different materials via Spherical Gaussians (SG), which, however, tends to blur the high-frequency reflection details.
In this paper, we propose an end-to-end inverse rendering pipeline that decomposes materials and illumination from multi-view images, while considering near-field indirect illumination.
In a nutshell, we introduce the Monte Carlo sampling based path tracing and cache the indirect illumination as neural radiance, enabling a physics-faithful and easy-to-optimize inverse rendering method.
To enhance efficiency and practicality, we leverage SG to represent the smooth environment illuminations and apply importance sampling techniques.
To supervise indirect illuminations from unobserved directions, we develop a novel radiance consistency constraint between implicit neural radiance and path tracing results of unobserved rays along with the joint optimization of materials and illuminations, thus significantly improving the decomposition performance.
Extensive experiments demonstrate that our method outperforms the state-of-the-art on multiple synthetic and real datasets, especially in terms of inter-reflection decomposition.

\end{abstract}

\section{Introduction}

Inverse rendering, \ie, recovering geometry, material and lighting from images, is a long-standing problem in computer vision and graphics. 
It is important for digitizing our real world and acquiring high quality 3D contents in many applications such as VR, AR and computer games.

\begin{figure}[tbp]
\centering
\includegraphics[width=\columnwidth]{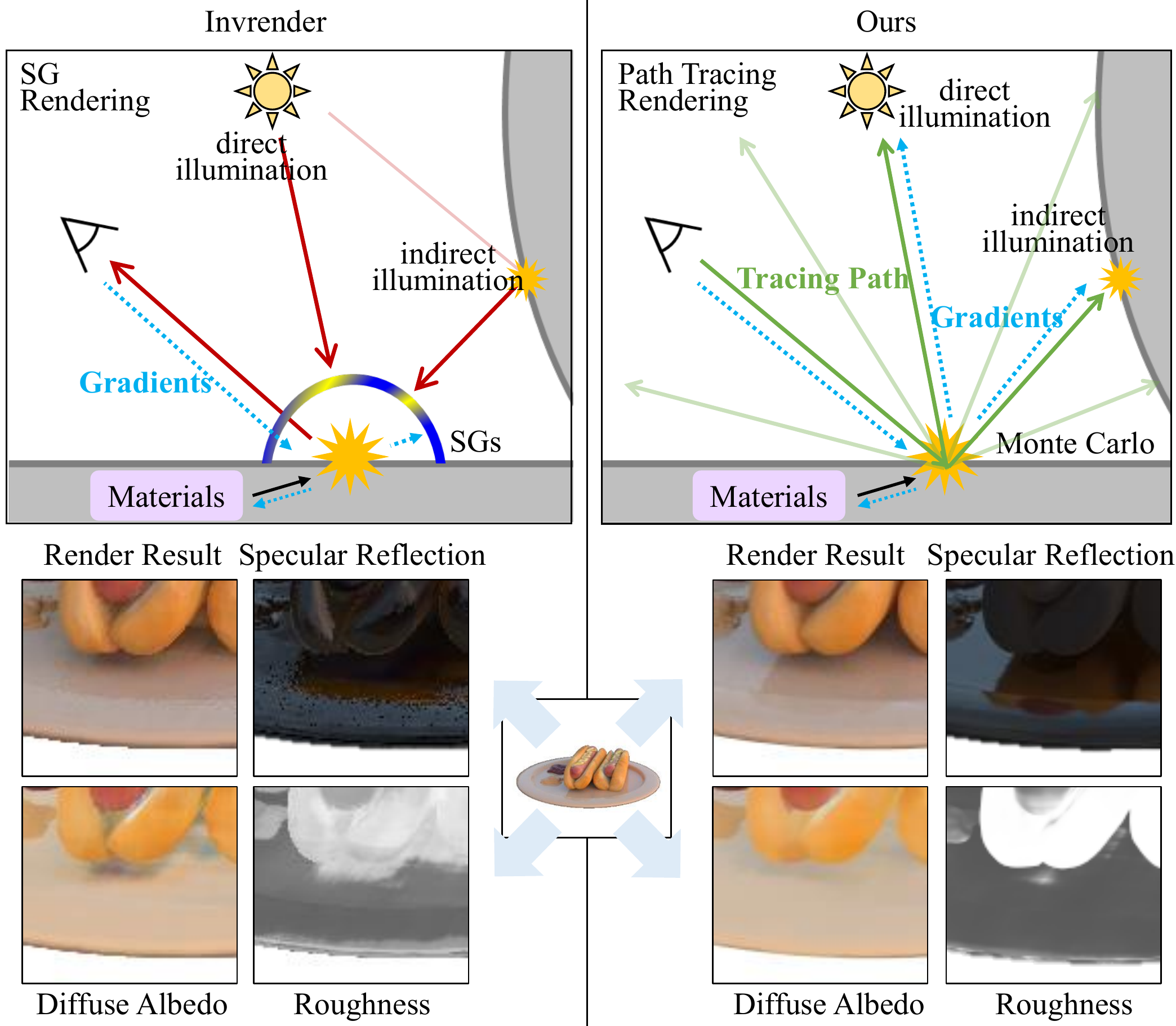}
\caption{
Our method integrates lights through path tracing with Monte Carlo sampling, while Invrender \cite{zhang2022modeling} uses Spherical Gaussians to approximate the overall illumination.
In this way, our method simultaneously optimizes indirect illuminations and materials, and achieves better decomposition of inter-reflections.
}
\label{fig:intro-main} 
\end{figure}

Recent methods \cite{zhang2021physg, zhang2021nerfactor, boss2021neural, zhang2022modeling} represent geometry and materials as neural implicit fields, and recover them in an analysis-by-synthesis manner. However, how to decompose the indirect illumination from materials is still challenging. Most methods \cite{boss2021nerd, zhang2021physg, zhang2021nerfactor, boss2021neural, munkberg2022extracting} model the environment illuminations but ignore indirect illuminations. As a result, the inter-reflections and shadows between objects are mistakenly treated as materials.
Invrender \cite{zhang2022modeling} takes the indirect illumination into consideration, and approximates it with Spherical Gaussian (SG) for computation efficiency. Since SG approximation cannot model the high frequency details, the recovered inter-reflections tend to be blurry and contain artifacts. 
Besides, indirect illuminations estimated by an SG network cannot be jointly optimized with materials and environment illuminations.

In this paper, we propose an end-to-end inverse rendering pipeline that decomposes materials and illumination, while considering near-field indirect illumination.
In contrast to the method \cite{zhang2022modeling}, we represent the materials and the indirect illuminations as neural implicit fields, and jointly optimize them with the environment illuminations.
Furthermore, we introduce a Monte Carlo sampling based path tracing to model the inter-reflections while leveraging SG to represent the smooth environment illuminations.
In the forward rendering, incoming rays are sampled and integrated by 
Monte Carlo estimator instead of being approximated by a pretrained SG approximator, as shown in \cref{fig:intro-main}.
To depict the radiance, the bounced secondary rays are further traced once and computed based on the cached neural indirect illumination.
During the joint optimization, the gradients could be directly propagated to revise the indirect illuminations.
In this way, high frequency details of the inter-reflection can be preserved.

Specifically, to make our proposed framework work, we need to address two critical techniques:

(i) The Monte Carlo estimator is computationally expensive due to the significant number of rays required for sampling. 
To overcome this, we use importance sampling to improve integral estimation efficiency. 
We also find that SG is a better representation of environment illuminations and adapt the corresponding importance sampling techniques to enhance efficiency and practicality.

(ii) Neural implicit fields often suffer generalization problems when the view directions deviate from the training views, which is the common case of indirect illumination. This would lead to erroneous decomposition between materials and illuminations. 
It is hard to determine whether radiance comes from material albedos or indirect illuminations as the indirect illuminations from unobserved directions are unconstrained or could have any radiance.
To learn indirect illuminations from unobserved directions, we introduce a radiance consistency constraint that enforces the implicit neural radiance produced by the neural implicit fields and path tracing results of unobserved directions. In this fashion, the ambiguity between materials and indirect illuminations has been significantly mitigated. Moreover, they can be jointly optimized with environment illuminations, leading to better decomposition performance.

We evaluate our method on synthetic and real data. 
Experiments show that our approach achieves better performance than others. 
Our method can render sharp inter-reflection and recover accurate roughness as well as diffuse albedo.
Our contributions are summarized as follows:
\begin{itemize}
\vspace{-0.6em}
    \item We propose an end-to-end inverse rendering pipeline that decomposes materials and illumination, while considering near-field indirect illumination.
    \vspace{-0.6em}
    \item We introduce the Monte Carlo sampling based path tracing and cache the indirect illumination as neural radiance, resulting in a physics-faithful and easy-to-optimize inverse rendering process.
    \vspace{-0.6em}
    \item We employ SG to parameterize smooth environment illumination and apply importance sampling techniques to enhance efficiency and practicality of the pipeline. 
    \vspace{-0.6em}
    \item We introduce a new radiance consistency in learning indirect illuminations, which can significantly alleviate the decomposition ambiguity between materials and indirect illuminations. 
\end{itemize}

\section{Related Work}

\subsection{Implicit Neural Representation}
Implicit neural representations~\cite{mildenhall2021nerf, yariv2020multiview, wang2021neus} have achieved impressive performance.
NeRF \cite{mildenhall2021nerf} represents scenes as radiance fields and volumetric density fields, and achieves photo-realistic novel view synthesis.
To better model geometry, some methods, such as IDR~\cite{yariv2020multiview} and NeuS~\cite{wang2021neus}, further represent geometry as Signed Distance Functions (SDFs).
However, the object appearance is represented as a radiance field, which simply outputs outgoing radiance of each 3D point given a view direction. Thus, the surface points can be treated as emissive lighting sources.
These methods are not suitable for relighting and material editing.

\begin{figure*}[tbp]
\centering
\includegraphics[width=1.0\textwidth]{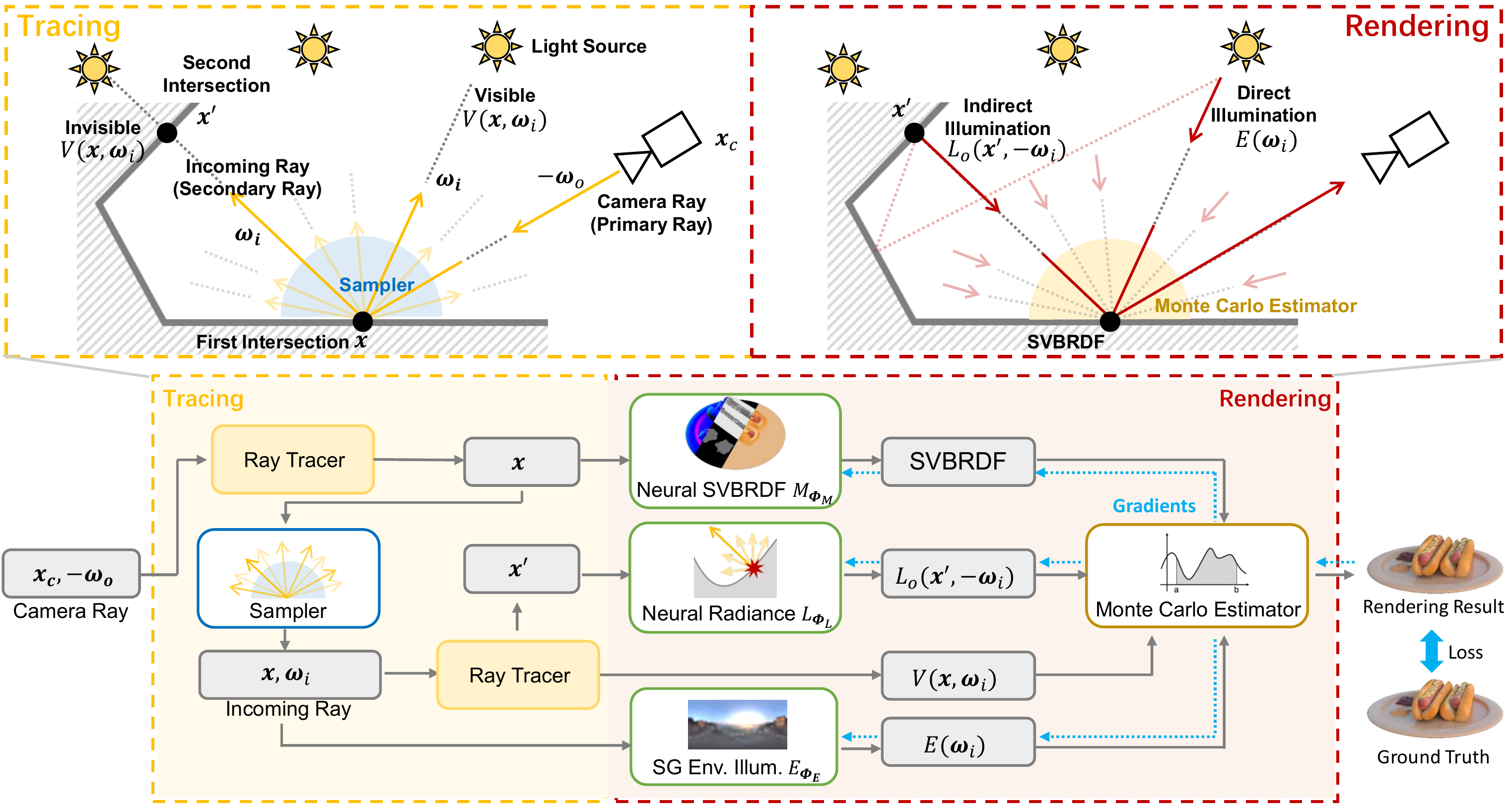}
\caption{
    \textbf{Proposed Rendering Pipeline.}
    To render a camera ray intersecting with surface at location $\boldsymbol{x}$, we first sample incoming rays and trace them to obtain their second surface intersection $\boldsymbol{x}'$ and visibility $V(\boldsymbol{x}, \boldsymbol{w}_i)$ for light source (environment illumination).
    Then, SVBRDF values at location $\boldsymbol{x}'$ and outgoing radiance $L_o(\boldsymbol{x}', -\boldsymbol{w}_i)$ of second intersection $\boldsymbol{x}'$, \ie, indirect illumination, are obtained by neural SVBRDF $M_{\boldsymbol{\Theta}_M}$ and neural radiance $L_{\boldsymbol{\Theta}_L}$, respectively.
    Besides, radiance of incoming rays from light source $E_{\boldsymbol{\Theta}_E}(\boldsymbol{w_i})$, \ie, direct illumination, is obtained by SG environment illumination $E_{\boldsymbol{\Theta}_E}$.
    Finally, a Monte Carlo Estimator is used for rendering the final results as described in \cref{eq:path_tracing_rendering}.
    Materials, indirect illumination and environment illumination are jointly optimized by the reconstruction loss.
}
\label{fig:method-framework} 
\end{figure*}

\subsection{Material and Illumination Estimation}
To estimate object materials, most of previous capture systems rely on constrained settings, such as by light-stages with controlled lights and cameras~\cite{guo2019relightables, lensch2003planned, zhang2021neural}, using moving cameras with co-located flashlight~\cite{bi2020neural, bi2020deep},  placing objects on a turntable platform, or capturing in special lighting patterns~\cite{kang2019learning}.
Apart from those hardware-specific systems, some data-driven methods~\cite{barron2014shape, li2020inverse, li2018learning, lichy2021shape, sang2020single, sengupta2019neural, wei2020object, yu2019inverserendernet} try to directly estimate materials from a single image by neural networks with priors from large-scale datasets.
However, they fail to generalize beyond the training datasets and are often restricted to the planar geometry.
Differentiable rendering methods~\cite{azinovic2019inverse, nimier2021material} aim to make graphic rendering process differentiable, and recover material and illumination by optimization.
However, they suffer from demanding computation cost and challenging optimization complexity.  

Recent works have been extended to more flexible capture settings by implicitly representing geometry and materials and optimizing them in differentiable pipelines.
Most methods adopt differentiable rendering algorithms and only consider direct illuminations, such as Spherical Gaussians (\eg, NeRD~\cite{boss2021nerd} and PhySG \cite{zhang2021physg}), Spherical Harmonics (\eg, NeROIC \cite{neroic2022kuang}), point lighting of low resolution environment maps (\eg, NeRFactor~\cite{zhang2021nerfactor}), and pre-filtered approximations (\eg, Neural-PIL \cite{boss2021neural} and NVDiffrec~\cite{munkberg2022extracting}).
Some methods \cite{hasselgren2022nvdiffrecmc,srinivasan2021nerv} integrated with Monte Carlo sampling but still ignore modeling multiple light bouncing, \eg, NVDiffrecmc~\cite{hasselgren2022nvdiffrecmc} only considers direct illumination and NeRV \cite{srinivasan2021nerv} considers only one indirect bounce.

Invrender \cite{zhang2022modeling}, most close to our method, considers multi-bounce indirect illuminations. It adopts SG rendering approximation and has to optimize in three stages.
Radiance field cannot be well trained with limited observed images at the first stage. Besides, incoming light of adjacent surface points may vary drastically because incoming light is represented as SGs and modeled by a coordinate-based network trained at the second stage.
In contrast, our method considers indirect illuminations and proposes a joint learning approach. Therefore, we can render sharp and complex self-reflection effects and recover material properties with higher quality, as shown in \cref{fig:intro-main}.

\subsection{Theoretical Rendering Process}
\label{sec:related-theoretical_rendering}
In theory, the rendering process at the intersection location $\boldsymbol{x}$ of the camera ray with direction $\boldsymbol{w}_o$ can be expressed by the rendering equation \cite{kajiya1986rendering}:
\begin{equation}
    \label{eq:rendering_equation}
    L_o(\boldsymbol{x}, \boldsymbol{w}_o) \!= \!\int_{\Omega}
            L_i(\boldsymbol{x}, \boldsymbol{w}_i)
            f_r(\boldsymbol{x}, \boldsymbol{w}_o, \boldsymbol{w}_i)
            (\boldsymbol{w}_i \cdot \boldsymbol{n}) 
        d\boldsymbol{w}_i,
\end{equation}
where $L_i(\boldsymbol{x}, \boldsymbol{w}_i)$ is the incoming radiance at surface point $\boldsymbol{x}$ along the direction $\boldsymbol{w}_i$, $f_r$ is the BRDF function and the outgoing radiance $L_o(\boldsymbol{x}, \boldsymbol{w}_o)$ in observed direction $\boldsymbol{w}_o$ is a reflected light integration over hemisphere $\Omega$ around the surface normal $\boldsymbol{n}$.
Incoming radiance may come directly form light source, known as direct illumination, or indirectly from other surface after multiple light bouncing, known as indirect illumination.
For indirect illumination, recursive rendering is often needed.

\section{Proposed Method}
\subsection{Overview}
Given a group of multi-view images captured under static illumination, we aim to decompose the geometry and Spatially Varying BRDF (SVBRDF) of the object and the illumination. 
We take the global illumination effect into consideration, such as shadows and inter-reflections, but consider transparent and translucent objects outside the scope of our work. 

The geometry is represented as the zero level set of SDF as in \cite{wang2021neus, yariv2020multiview, zhang2022modeling}, which is modeled by an MLP that maps a 3D location $\boldsymbol{x} \in \mathbb{R}^{3}$ to an SDF value and a geometric feature vector $\boldsymbol{f} \in \mathbb{R}^{512}$.
The material is encoded by another MLP as neural SVBRDF $M_{\boldsymbol{\Theta}_M}(\boldsymbol{x},  \boldsymbol{f})$.
The environment illumination is parameterized by SG coefficients \cite{zhang2021physg} $E_{\boldsymbol{\Theta}_E}(\boldsymbol{w_i})$ , where $\boldsymbol{w_i} \in \mathbb{R}^{2}$ is the light direction.
The radiance is represented as another MLP $L_{\boldsymbol{\Theta}_L}(\boldsymbol{x},  \boldsymbol{n}, \boldsymbol{w}_o, \boldsymbol{f})$, which outputs outgoing radiance $L$ given location $\boldsymbol{x}$, normal $\boldsymbol{n} \in \mathbb{R}^{3}$, viewing direction $\boldsymbol{w}_o \in \mathbb{R}^{3}$ and feature $\boldsymbol{f}$.

We solve the inverse rendering problem in an analysis-by-synthesis manner by forward rendering with parameterized components.
Similar to prior works, we pretrain the geometry SDF by NeuS \cite{wang2021neus} and freeze the parameters.
Given a viewing direction $\boldsymbol{w_o}$, we first find the intersection $\boldsymbol{x}$ on the geometry surface through sphere tracing on the SDF. 
Then, the path tracing based rendering integrates the outgoing radiance $L_o(\boldsymbol{x}, \boldsymbol{w}_o)$ in direction $\boldsymbol{w_o}$.
The rendering results are compared with the input image pixels to optimize $\boldsymbol{\Theta}_L$, $\boldsymbol{\Theta}_M$ and $\boldsymbol{\Theta}_E$.

\subsection{Cached Path Tracing based Rendering}
\label{sec:major_forward_rendering}

Theoretical rendering process described in \cref{sec:related-theoretical_rendering} cannot be practically implemented because of its production integration and exponential recursive light bounce.
In contrast to simply ignoring light bounce and adopting approximations rendering method such as SG \cite{wang2009all}, we implement the forward rendering process based on path tracing~\cite{lafortune1996mathematical, lafortune1993bi}, which is an efficient and differentiable rendering framework that fully incorporates the light bounces.
We implement rendering equation in \cref{eq:rendering_equation} by Monte Carlo estimator as:
\begin{equation}
    \label{eq:path_tracing_rendering}
    L_o(\boldsymbol{x}, \boldsymbol{w}_o) \approx 
        \frac{1}{N} 
        \sum_{i=1}^N 
        \frac{
            L_i(\boldsymbol{x}, \boldsymbol{w}_i)
            f_r(\boldsymbol{x}, \boldsymbol{w}_o, \boldsymbol{w}_i)
            (\boldsymbol{w}_i \cdot \boldsymbol{n})
        } 
        {p(\boldsymbol{w}_i)} 
        .
\end{equation}
It estimates the production integration by sampling incoming rays with direction $\boldsymbol{w}_i$ drawn from distribution $p(\boldsymbol{w}_i)$. 

The incoming radiance includes light rays directly emitted by the light source, \ie, direct illumination, and ones bouncing off of the object surface multiple times, \ie, indirect illumination:
\begin{equation}
    \label{eq:light_bounce}
    L_i(\boldsymbol{x}, \boldsymbol{w}_i) = 
        V(\boldsymbol{x}, \boldsymbol{w}_i) E(\boldsymbol{w}_i) 
        + (1 - V(\boldsymbol{x}, \boldsymbol{w}_i))L_o(\boldsymbol{x}', -\boldsymbol{w}_i)
        ,
\end{equation}
where $E(\boldsymbol{w}_i)$ is the incoming radiance form light source along direction $\boldsymbol{w}_i$, and $L_o(\boldsymbol{x}', -\boldsymbol{w}_i)$ is the incoming radiance from the second intersection $\boldsymbol{x}'$ of the ray. 
$V(\boldsymbol{x}, \boldsymbol{w}_i)$ is the visibility of location $\boldsymbol{x}$ for light source and indicates the illumination type, obtained during path tracing.

To obtain indirect illumination, in theory, we should recursively render the outgoing radiance at location $\boldsymbol{x}'$ along direction $-\boldsymbol{w}_i$ by \cref{eq:rendering_equation}.
This may lead to intractable computation and optimization difficulties for the optimization process.
Inspired by \cite{zhang2022modeling}, we employ the neural radiance $L_{\boldsymbol{\Theta}_L}$ to represent the final outgoing radiance after multiple light bouncing of the second ray intersection $\boldsymbol{x}'$, known as indirect illumination.
In such manner, we cache the indirect illumination and avoid the exhaustive ray tracing. The indirect incoming radiance 
is calculated as:
\begin{equation}
    L_o(\boldsymbol{x}', -\boldsymbol{w}_i) = 
    L_{\boldsymbol{\Theta}_L}(\boldsymbol{x}', \boldsymbol{n}', -\boldsymbol{w}_i, \boldsymbol{f}'),
\end{equation}
where $\boldsymbol{n}'$ and $\boldsymbol{f}'$ are the surface normal and geometric feature vector at $\boldsymbol{x}'$ respectively.

The complete pipeline of our path tracing based rendering is shown in \cref{fig:method-framework}.
The rendering process is differentiable for optimizing neural radiance $L_{\boldsymbol{\Theta}_L}$, neural SVBRDF $M_{\boldsymbol{\Theta}_M}$ and SG environment illumination  $E_{\boldsymbol{\Theta}_E}$.

\subsection{Efficient Monte Carlo Estimator}
Monte Carlo estimator needs to sample a large number of rays to produce high-quality results without noise, which is not affordable for practical optimization.
Although some techniques can tackle the issue, most of them are inappropriate in inverse rendering scenario.
For example, denoising techniques \cite{recentadvances2015zwicker, bako2017kernel, chaitanya2017interactive, hasselgren2020neural} require spatial information of the whole rendered image and temporal information from previous frames.
These information is not available in inverse rendering, where we randomly pick posed images and sample some pixels for training. 
Hence, we apply importance sampling techniques, including cosine sampling as well as GGX importance sampling \cite{heitz2018sampling}, to improve Monte Carlo estimator efficiency and use multiple importance sampling method \cite{veach1995optimally, pharr2016physically} to fuse all of them.

For light importance sampling, the piecewise-constant 2D distribution sampling\cite{pharr2016physically} is not applicable, since it is designed for known environment illumination represented as the 2D array.
As mentioned in \cite{zhang2022modeling, boss2021nerd}, parameterizing environment illumination in such a way could make each pixel of environment maps vary independently, lead diffuse albedo baked in illumination and cause illumination inefficient for optimization.
In contrast, we parameterize environment illumination as SG coefficients 
, and introduce and adapt Spherical Gaussian (SG) distribution sampling~\cite{jakob2012numerically} as the corresponding light importance sampling technique:

\begin{equation}
    p_{SG}(\boldsymbol{w}_i) = 
    \sum_{k=1}^M 
    a_k 
    \frac{\lambda_k}{2\pi(1-e^{-2\lambda_k})}
    e^{\lambda_k(\boldsymbol{w}_i \cdot \boldsymbol{\xi}_k - 1)}
    ,
\end{equation}
\begin{equation}
    \label{eq:importance_weight}
    a_k = 
    \frac{\bar{\mu}_k max(\boldsymbol{n} \cdot \boldsymbol{\xi}_k, \epsilon)}
    {\sum_{j=1}^M \bar{\mu}_j max(\boldsymbol{n} \cdot \boldsymbol{\xi}_j, \epsilon)},
\end{equation}
where $\boldsymbol{\xi}, \lambda, \boldsymbol{\mu} \in \boldsymbol{\Theta}_E$ are SG parameters of environment illumination, \ie, lobe axis, lobe sharpness and lobe amplitude of SG respectively, and $\bar{\mu}$ is the energy of lobe amplitude $\boldsymbol{\mu}$.
Since we only need to sample rays over the hemisphere around $\boldsymbol{n}$, we assign a tiny weight $\epsilon$ to SG components whose lobe axis $\boldsymbol{\xi}$ is beyond the hemisphere. 
According the SG distribution, it has a higher probability to sample light rays that belong to brighter SG lobes and are closer to SG lobe centers.
The 
detailed 
process is described in our supplemental material.

\subsection{Training with Traced Rays}
\label{sec:secondary_ray_training}
We alternatively train our framework with observed rays and unobserved rays.
Training with observed rays alone is challenging because some locations or view directions are not observed due to occlusion. %
Besides, there exists ambiguity between indirect illumination and material properties, since indirect incoming rays with many directions cannot be directly observed by the camera.
Hence, neural radiance $L_{\boldsymbol{\Theta}_L}$ is indeterminate with re-render loss alone.
We propose to utilize the unobserved rays to provide more information and constraints.

\begin{figure}[tbp]
\centering
\includegraphics[width=\columnwidth]{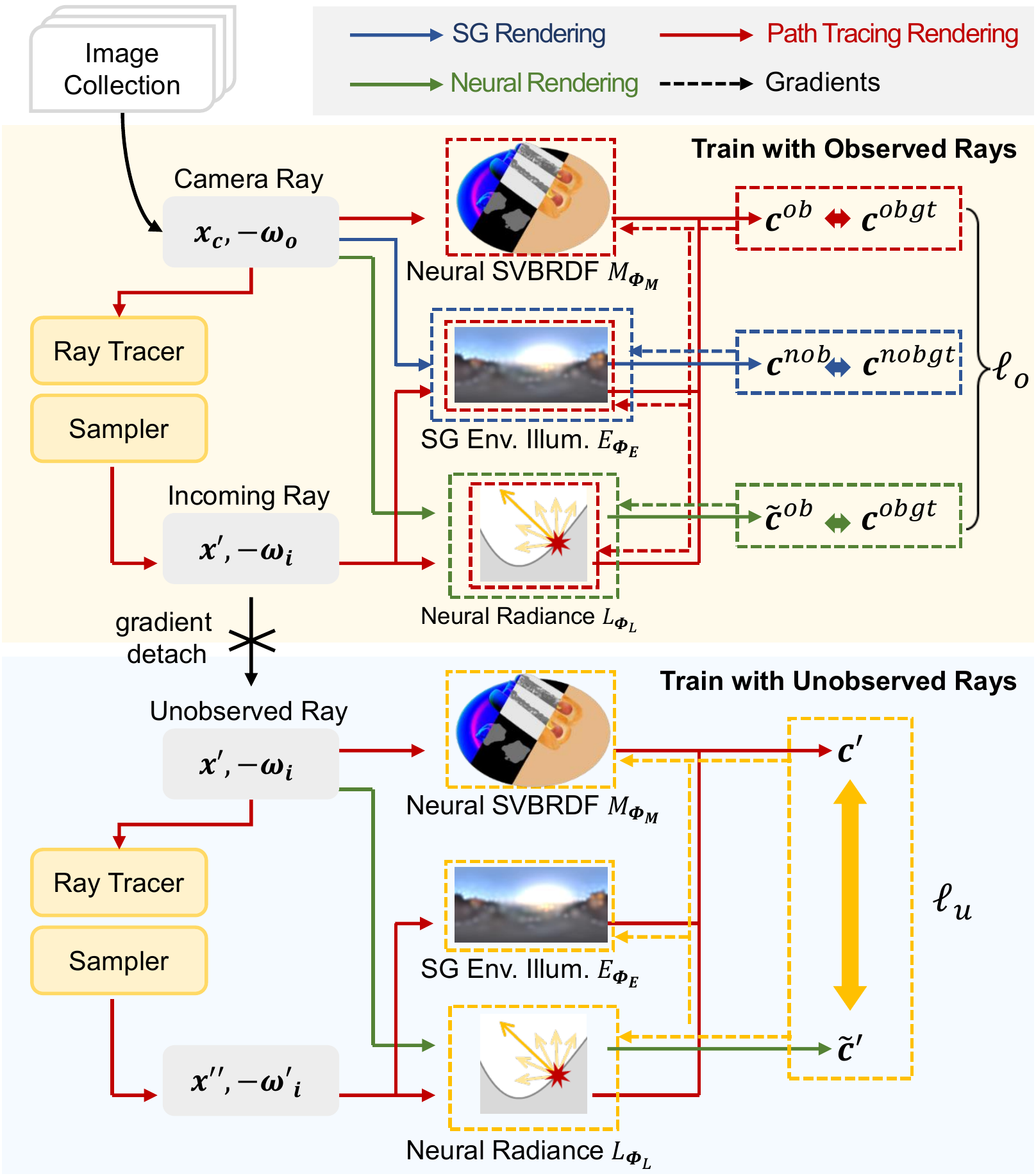}
\caption{
    \textbf{Training with traced rays.}
    We alternatively train with observed rays and unobserved rays.
}
\label{fig:method-training} 
\end{figure}

\noindent\textbf{Train with observed rays.}
As shown in the top of \cref{fig:method-training}, we optimize $\boldsymbol{\Theta}_L$, $\boldsymbol{\Theta}_M$ and $\boldsymbol{\Theta}_E$ with observed rays using the following loss:
\begin{equation}
\label{eq:loss_primary}
\begin{split}
    {\ell}_{o} = 
        & \frac{1}{N_{obj}}\sum_{i=1}^{N_{obj}} \| \boldsymbol{c}_i^{ob} - \boldsymbol{c}_i^{obgt} \|_1  \\
        & + \beta_1 \frac{1}{N_{obj}}\sum_{i=1}^{N_{obj}} \| \boldsymbol{\widetilde c}_i^{ob} - \boldsymbol{c}_i^{obgt} \|_1  \\
		& + \beta_2 \frac{1}{N_{nobj}}\sum_{i=1}^{N_{nobj}} \| \boldsymbol{c}_i^{nob} - \boldsymbol{c}_i^{nobgt} \|_2
     .
\end{split}
\end{equation}

The first term is the reconstruction loss of path-tracing-based rendering results of object pixels $\{\boldsymbol{c}_i^{ob}\}_{i=1}^{N_{obj}}$ with ground truth $\{\boldsymbol{c}_i^{obgt}\}_{i=1}^{N_{obj}}$.
The second term is the reconstruction loss of neural rendering results of object pixels $\{\boldsymbol{\widetilde c}_i^{ob}\}_{i=1}^{N_{obj}}$.
The third term is the environment reconstruction loss, which renders non-object pixels $\{\boldsymbol{c}_i^{nob}\}_{i=1}^{N_{nobj}}$ to compare with the ground truth $\{\boldsymbol{c}_i^{nobgt}\}_{i=1}^{N_{nobj}}$.

\noindent\textbf{Train with unobserved rays.}
As shown in the bottom of \cref{fig:method-training}, we additionally optimize components with unobserved rays.
Although there is no ground truth of unobserved rays, the consistency of neural rendering and path tracing based rendering of the rays can be used for training:
\begin{equation}
    \label{eq:loss_secondary}
    {\ell}_{u} = 
    \frac{1}{N_{sec}}\sum_{j=1}^{N_{sec}}
    \| {\boldsymbol{c}'}_j - {\boldsymbol{\widetilde c}'}_j \|_1
    ,
\end{equation}
where ${\boldsymbol{c}'}_j = L_o(\boldsymbol{x}', -\boldsymbol{w}_i)$ is the path tracing rendering result at the unobserved ray origin $\boldsymbol{x}'$ for outgoing direction $-\boldsymbol{w}_i$, 
and ${\boldsymbol{\widetilde c}'}_j = L_{\boldsymbol{\Theta}_L}(\boldsymbol{x}', \boldsymbol{n}', -\boldsymbol{w}_i, \boldsymbol{f}')$ is the neural rendering result.
$N_{sec}$ is the amount of unobserved rays.

Unobserved rays are uniformly sampled from the secondary rays, which are generated in path tracing of observed rays, instead of being generated by virtual cameras. 
We alternatively train the networks with observed rays and unobserved rays rather than aggregate the two losses.
The unobserved rays is optimized every $K$ steps.

\begin{figure*}[tbp]
\centering
\includegraphics[width=\textwidth]{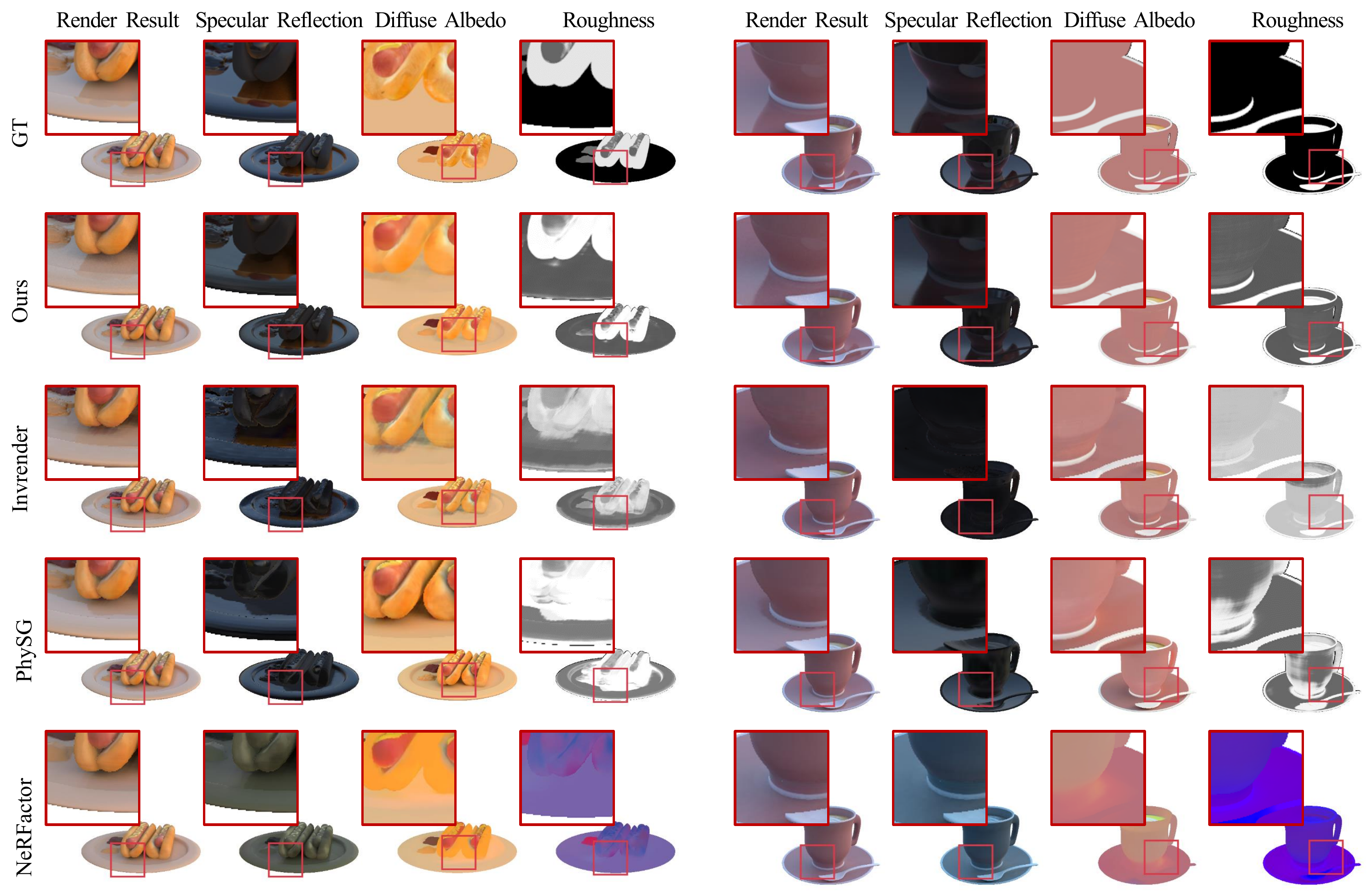}
\caption{
    \textbf{Qualitative comparisons with the state-of-the-art.}
    We present synthetic rendering result and specualar reflection component as well as estimated aligned diffuse albedo~\cite{zhang2021physg, zhang2021nerfactor} and roughness of each method on two scenes. 
    The roughness of NerFactor\cite{zhang2021nerfactor} is visualized with the BRDF identity latent code.
    Compared with previous works, our method better simulates sharp self-reflectance and separates shadows and indirect illumination from diffuse albedo. Besides, roughness maps recovered by our method are more accurate.
}
\label{fig:exp-camparison} 
\end{figure*}

\subsection{Implementation Details}

We set the sampled number of rays $N=64$ and the SG components number $M=128$.
We set the loss weights $\beta_1 = 1.0$, $\beta_2=1.0$ and unobserved rays training interval $K = 10$.
SDF and neural SVBRDF contains 8 layers with 512 hidden units and positional encoding \cite{tancik2020fourier, mildenhall2021nerf} is applied to the input 3D locations with 6 and 10 frequency components respectively.
Neural radiance contains 4 layers with 512 hidden units and positional encoding of the 3D location and directions with 10 and 4 frequency components respectively.
Our approach is implemented in Pytorch~\cite{paszke2019pytorch} and optimized with Adam with learning rate $5 \times 10^{-4}$.
We train about 120 epochs on 4 RTX 3090 GPUs and it takes about 5 hours.
We use the simplified Disney BRDF model~\cite{burley2012physically} with parameters including roughness, diffuse albedo and specular albedo.
The specular albedo is assumed as 0.5, the value of common dielectric surfaces.%
For stable optimization, we fix roughness at the first 50 epochs to warm up.

\section{Experiment}
\subsection{Synthetic Data}
We collect four synthetic scenes with obvious self-reflections to showcase the quality of the estimated BRDF parameters and illumination.
We render 200 images and their masks under a natural HDRI environment map via Blender Cycles and uniformly sample 100 for training and leave the rest for testing.
We also render diffuse albedo maps, roughness maps, and specular reflection components for test images to evaluate the inverse rendering ability.
The image resolution is set to 512 $\times$ 512.

\begin{table*}[tbp]
\centering
\renewcommand\arraystretch{1.3}
\scalebox{0.9}{
    \input{./tab-baseline_comparison.tex}
}
\caption{
    \textbf{Quantitative evaluations.}
    We present quantitative comparison with the state-of-the-art.
    Results show that our method achieves impressive improvements, especially in roughness estimation and specular reflection synthesis.
    Due to the rendering noise of our path tracing based rendering, some metrics of the view synthesis RGB are slightly worse than other methods. 
}
\label{tab:baseline_comparison}
\end{table*}

\subsection{Comparison with the State-of-the-Art}
The closest work to ours is Invrender\cite{zhang2022modeling} which forms our primary comparisons.
We also compare with other methods tackling on the similar inverse rendering settings as this paper for thorough comparisons, including NeRFactor \cite{zhang2021nerfactor} and PhySG \cite{zhang2021physg}.
We mainly focus the evaluation on material properties and illumination estimation instead of shape reconstruction.
We make quantitative comparisons on the synthetic data and directly learn geometry from mesh for every approach to better evaluate material estimation ability without interference of geometry reconstruction quality.
Following previous works \cite{zhang2022modeling, zhang2021physg}, we adopt Peak Signal-to-Noise Ratio (PSNR), Structural Similarity Index Measure (SSIM)~\cite{wang2004image}, and Learned Perceptual Image Patch Similarity (LPIPS)~\cite{zhang2018unreasonable} to evaluate image quality metrics and evaluate the diffuse albedo after aligning.

\cref{fig:exp-camparison} shows that our method could render sharp reflection effect due to our joint-learning path-tracing-based framework.
Hence, our recovered diffuse albedo and roughness are more clean compared with other methods.
Besides, by joint-learning framework, our indirect illumination and visibility are modeled more accurately, so less indirect illumination and shadow are baked into diffuse albedo.
\cref{tab:baseline_comparison} shows quantity improvements, especially in roughness estimation and specular reflection synthesis, of our methods.

Invrender\cite{zhang2022modeling} approximates the indirect illumination with SG and is trained in three stages.
They represent visibility and incoming indirect light of each point as SG parameters by neural networks and train in the second stage, then optimize materials with SG rendering at the third stage.
SG does not work well for high-frequency lighting, and the visibility and indirect illumination of adjacent surface points may vary drastically, hence, reflection tend to be noisy and rough, as shown in rendering RGB and specular RGB results in \cref{fig:exp-camparison}.
Besides, the radiance field, trained with limited observed rays of multi-view images, could not predict radiance of indirect rays with unobserved directions correctly.
Hence, more indirect illumination and shadow are baked in diffuse albedo as shown in \cref{fig:exp-camparison}.

Other methods\cite{zhang2021physg, zhang2021nerfactor} ignore indirect illumination and achieve worse results of material recovering.
Indirect illumination is baked in diffuse albedo and roughness maps are recovered inaccurately.

\begin{figure}[tbp]
\centering
    \includegraphics[width=\columnwidth]{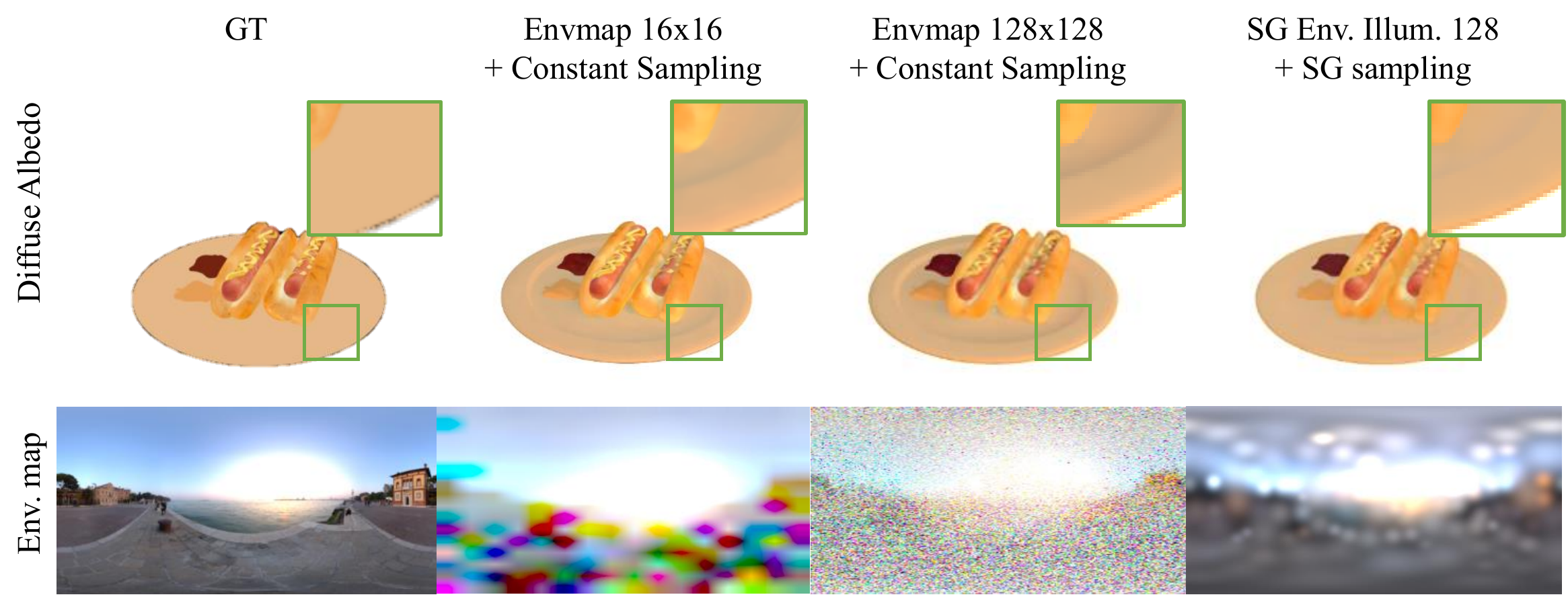}
    \label{fig:exp-sg_b}
    \caption{
        \textbf{Ablation on environment illumination representation.}
        Representing environment illumination in environment maps and using 2D piece-constant sampling causes neighbor pixels in the environment map vary independently, and part of the illumination is baked into diffuse albedo.
        Representing environment illumination in SG coefficients and using SG importance sampling better decomposes the illumination and diffuse albedo. 
    } 
\end{figure}

\begin{figure}[tbp]
\centering
\includegraphics[width=0.9\columnwidth]{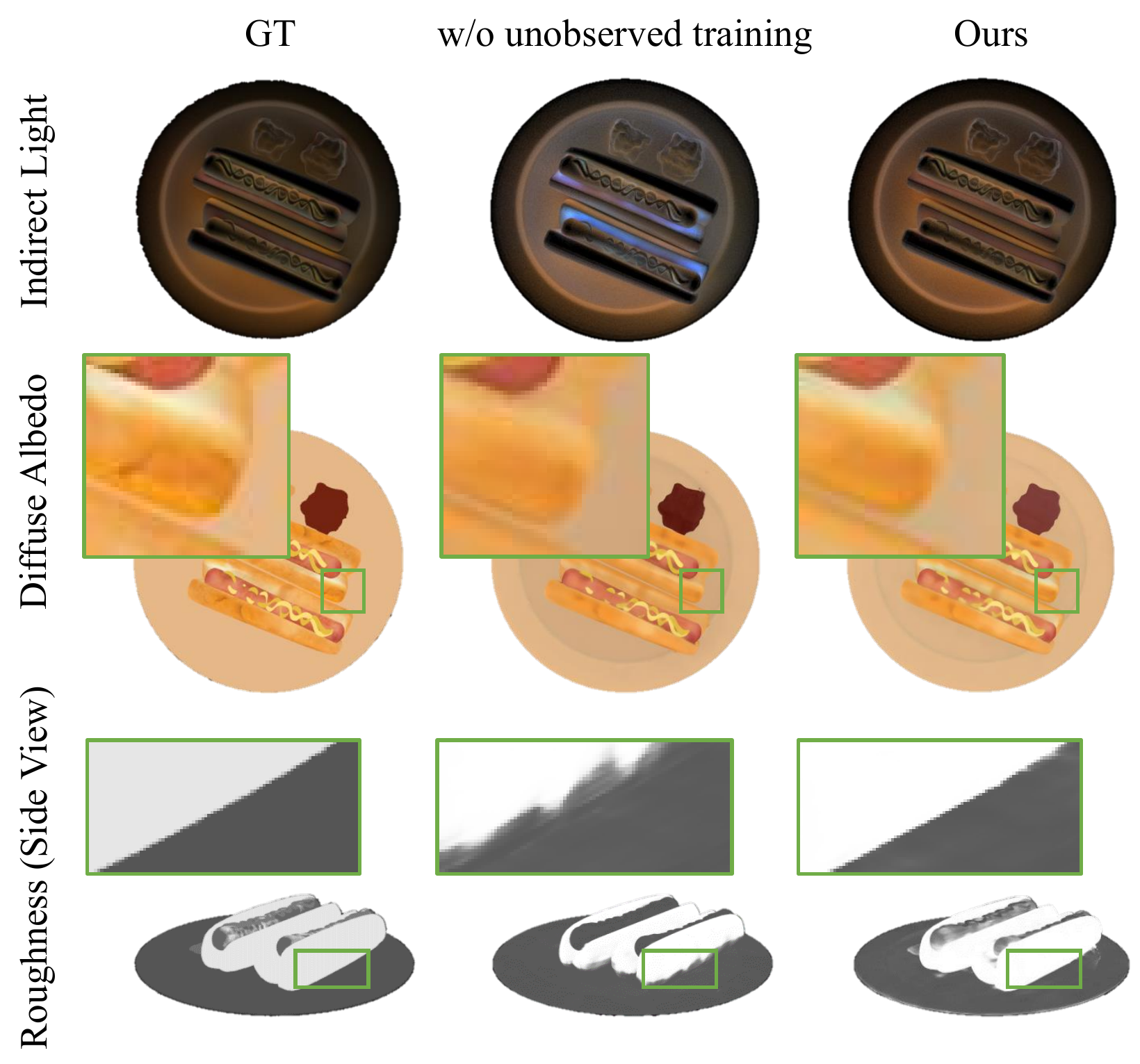}
\caption{
    \textbf{Ablations on training with unobserved rays.}
    We visualize the incoming indirect light for each point and present recovered diffuse albedo and roughness under both settings.
    Training with unobserved rays helps the decomposition of indirect light and diffuse albedo.
    Besides, roughness at the interstice of objects is recovered more accurately.
}
\label{fig:exp-ablation_secondary} 
\end{figure}

\subsection{Ablation Studies}

\noindent\textbf{Ablations on environment illumination representation.} 
As shown in \cref{fig:exp-sg_b}, representing environment illumination in SG coefficients and using SG importance sampling are better for optimization.
Representing environment illumination in 2D environment maps and using piece-constant sampling causes neighbor pixels in the environment map vary independently, and it is easier for illumination to be baked into diffuse albedo.

\noindent\textbf{Ablations on training with unobserved rays.} 
We ablate the unobserved training, and compare the results in \cref{fig:exp-ablation_secondary}.
We visualize the mean of indirect illumination from all directions at each point and show the recovered diffuse albedo as well as roughness.
Without unobserved training, the network predicts wrong indirect illumination at some locations, especially at the interstice of objects.
Due to the incorrect indirect illumination, the recovered diffuse albedo contains some artifacts, \eg, the bread on the side of the sausage of the hotdog.
Besides, interstices, \eg, areas between the hotdog and the plane, are not visible by cameras from many directions. Hence, the roughness at these areas cannot be estimated correctly and confidently when only trained with observed rays.

\begin{figure}[tbp]
\centering
\includegraphics[width=0.95\columnwidth]{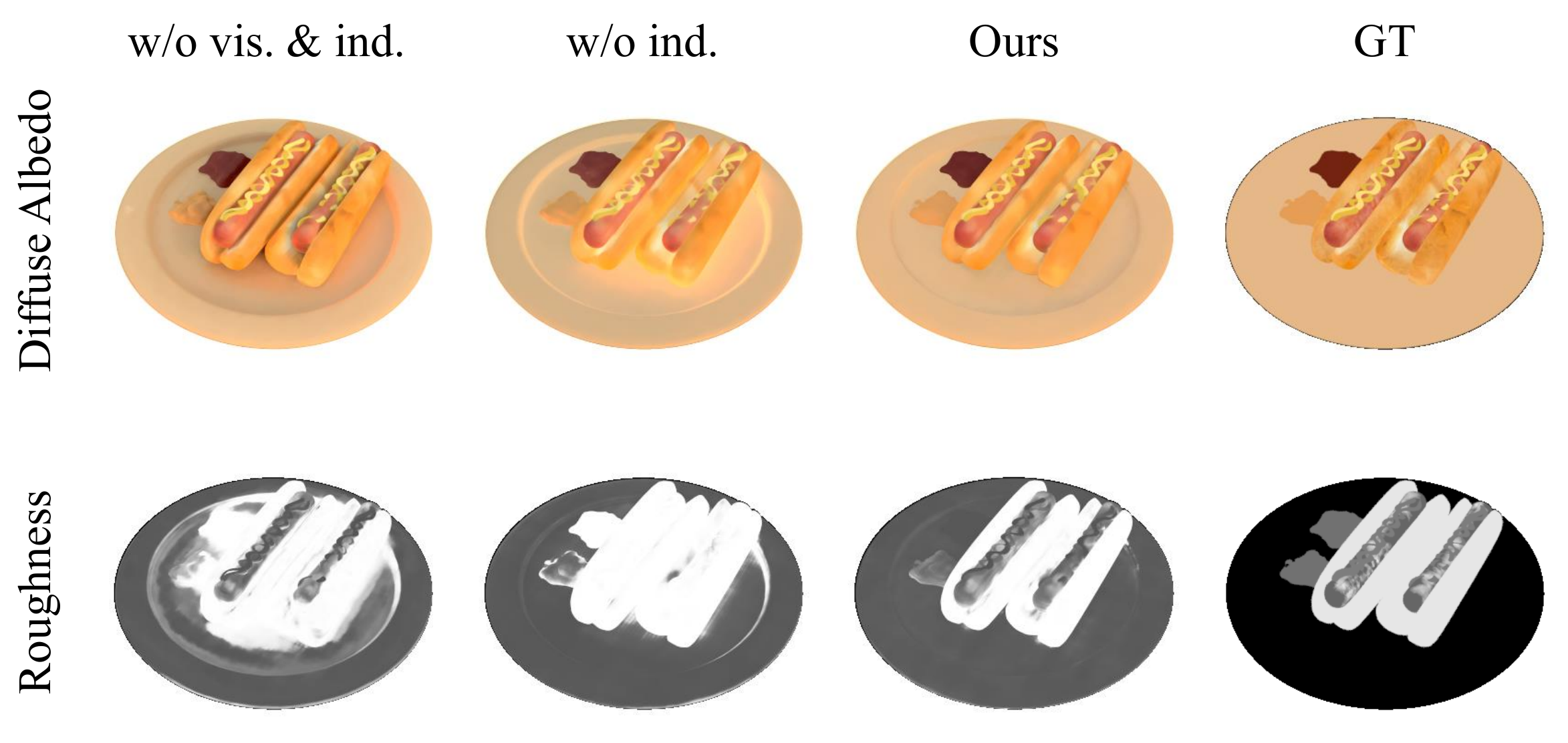}
\caption{
    \textbf{Ablation on indirect lighting.}
    Without modeling indirect illumination and visibility, indirect illumination and shadows would be baked into diffuse albedo and roughness.
}
\label{fig:exp-indirect} 
\end{figure}

\noindent\textbf{Ablations on indirect lighting.} 
We show the influence of modeling indirect illumination and visibility in \cref{fig:exp-indirect}.
Without modeling indirect illumination, indirect illumination would be baked into diffuse albedo, resulting in wrong brightness.
Further without modeling visibility, shadows would be baked into diffuse albedo and the roughness is not correctly recovered.
These results show the necessity of indirect illumination modeling in inverse rendering.

\subsection{Relighting}
\begin{figure}[tbp]
\centering
\includegraphics[width=0.95\columnwidth]{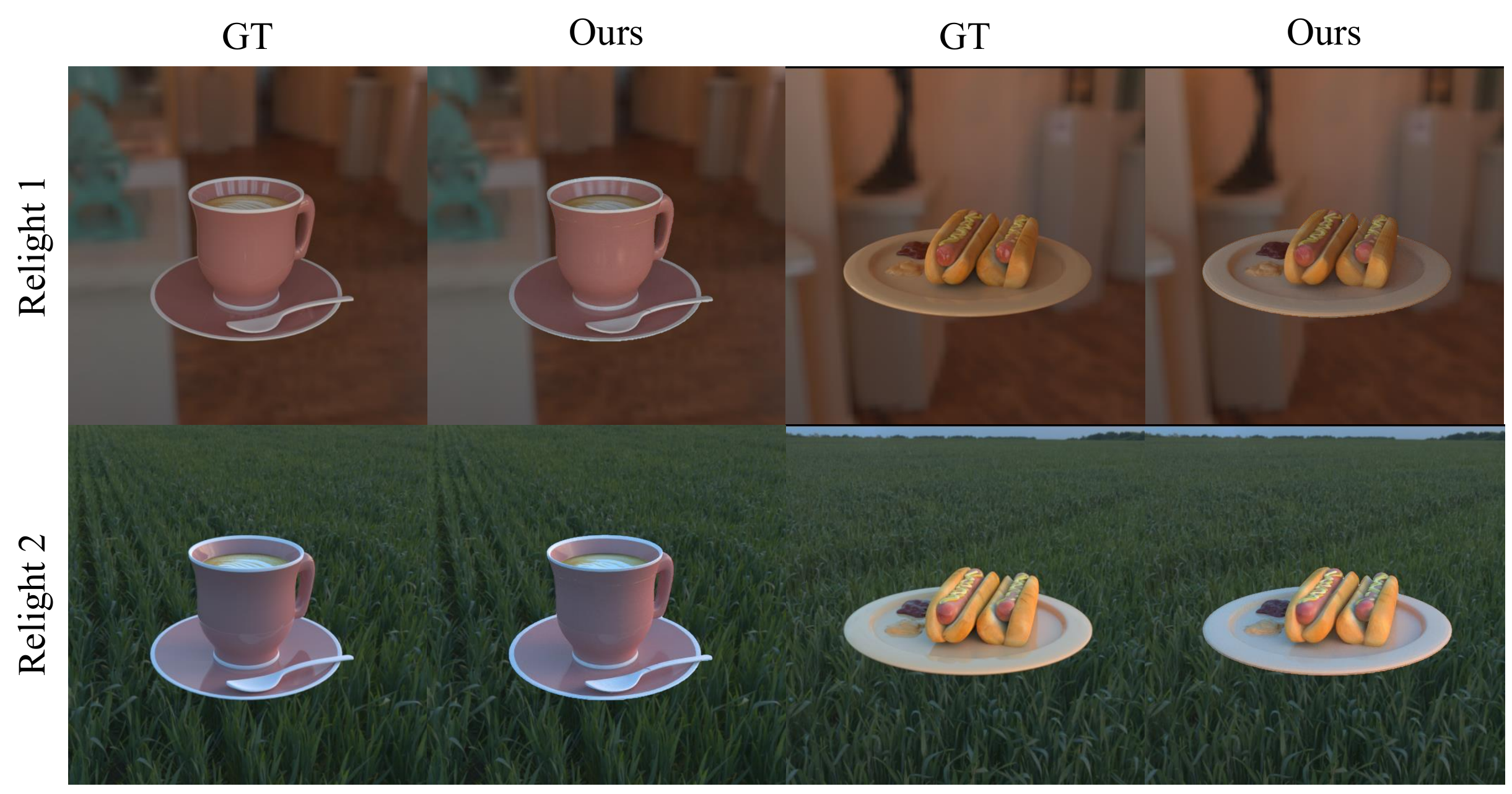}
\caption{
    \textbf{Relighting Results.}
    Our method supports further relighting with the recovered materials.
}
\label{fig:exp-relight} 
\end{figure}

We relight the objects with recovered material properties under two environment illuminations and show results in  \cref{fig:exp-relight}.
Our method could recover accurate material properties and support further relighting.

\subsection{Results on Real Captures}
\begin{figure}[tbp]
\centering
\includegraphics[width=0.98\columnwidth]{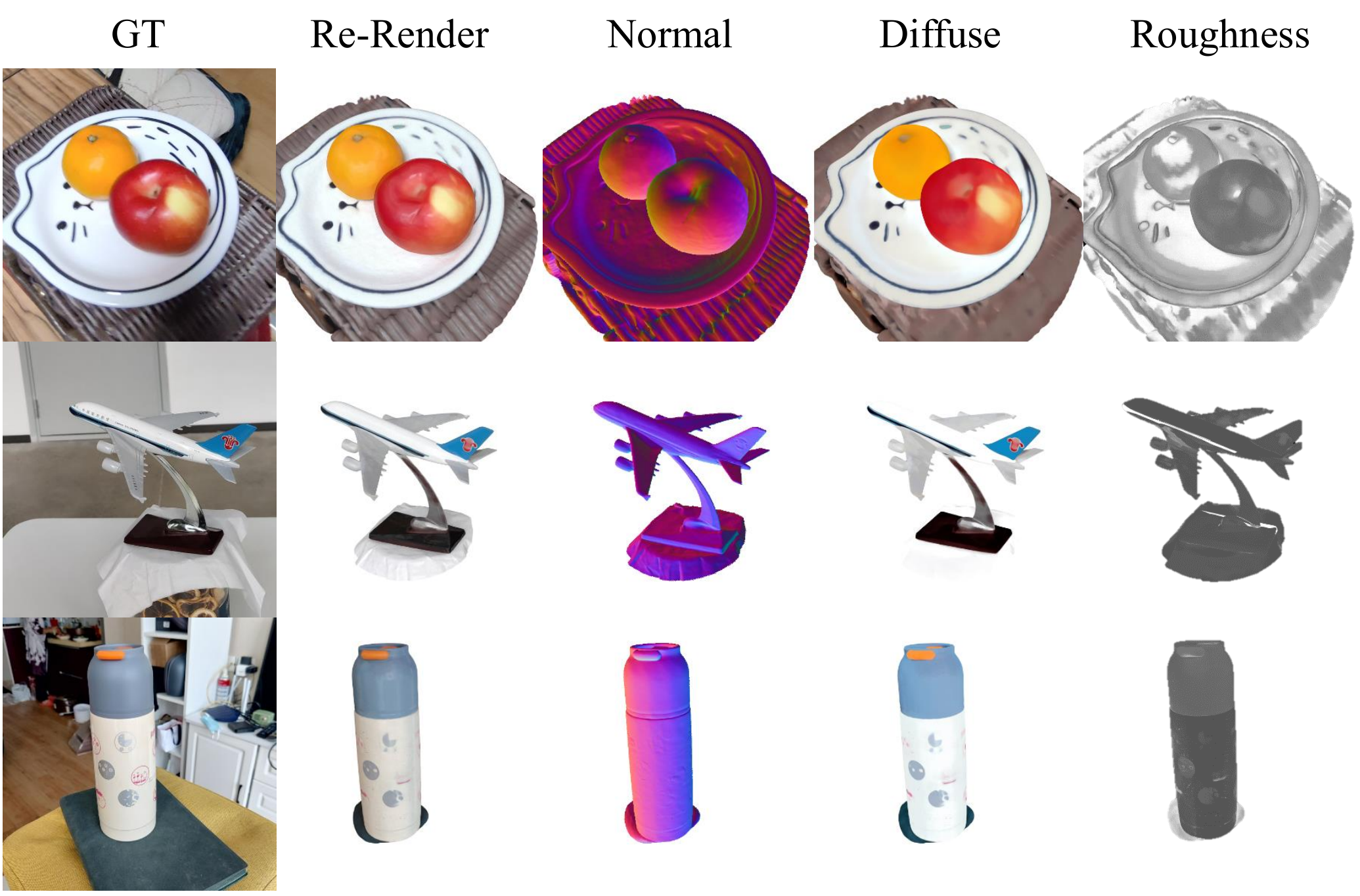}
\caption{
    \textbf{Results on real captures.}
    Our method estimates reasonable materials for real-world objects.
}
\label{fig:exp-real} 
\end{figure}

We test our method on real captured images of 3 objects with different materials.
Each scene has about 40 to 60 valid images for training and we use COLOMAP \cite{schonberger2016structure} to estimate the camera poses.
We train our method without masks.
Note that reflection properties of real materials are more complex in contrast to BRDF models and there are more interference in real capturing, \eg, video motion blur caused by moving cameras and illumination changing during capturing.
As shown in \cref{fig:exp-real}, our method could estimate reasonable material properties.

\begin{figure}[tbp]
\centering
\includegraphics[width=0.98\columnwidth]{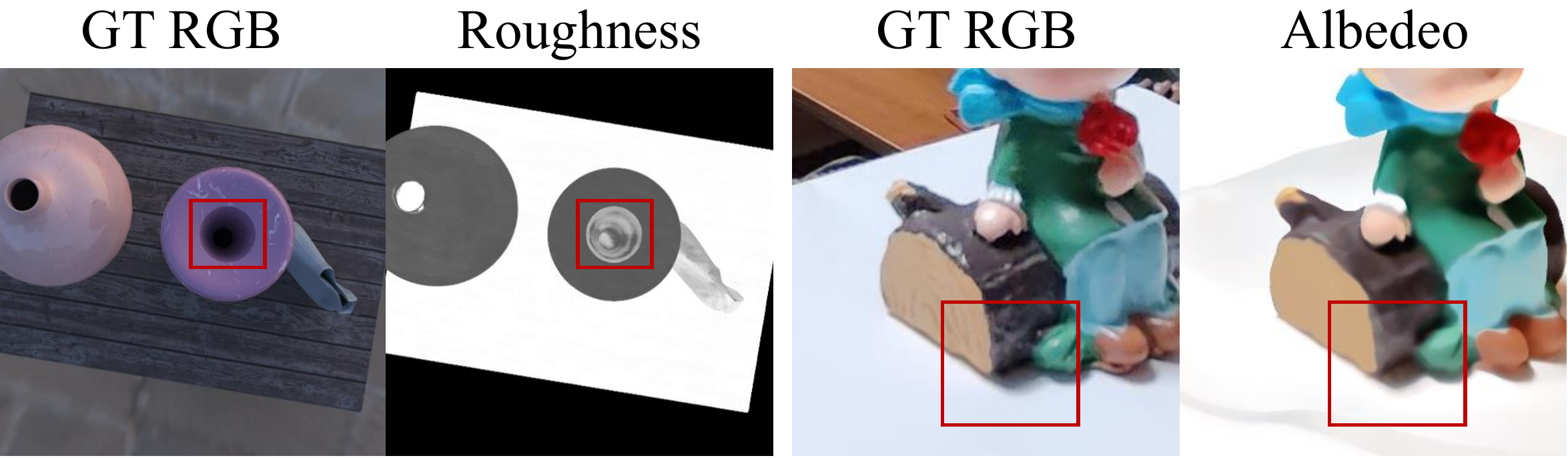}\\
\caption{\label{fig:failure} \textbf{Failure cases.}
Shadow might pose challenges for reflectance decomposition in some extreme cases.
}
\end{figure}

\subsection{Failure Cases}
As shown in \cref{fig:failure}, our method has difficulty in estimating roughness in large shadow areas due to the low visibility of scenes. In some extreme cases, the shadow may leak into the albedo because of illumination ambiguity.

\section{Conclusion}
To summarize, our paper presents an end-to-end inverse rendering pipeline  that is capable of decomposing materials and illumination from multi-view images, while considering near-field indirect illumination. 
Our method utilizes the Monte Carlo sampling based path tracing and cache the indirect illumination as neural radiance, enabling a physics-faithful and easy-to-optimize inverse rendering method.
We implement an efficient Monte Carlo estimator and propose a novel radiance consistency constraint of unobserved rays to decrease the ambiguity.
Extensive experiments demonstrate that our method models the sharp inter-reflections better and recovers material properties more accurately.

Our method still has some limitations.
First, the shape is not joint optimized because visibility gradients are not handled well by current ray tracing technique.
Second, to decrease the ambiguity of the inverse problem, the specular albedo is assumed as 0.5, the value of common dielectric surfaces.
They will be the subject of our future works.

\vspace{0.5em}
\noindent{\textbf{Acknowledgements.}} This research is funded in part by ARC-Discovery grant (DP220100800 to XY) and ARC-DECRA grant (DE230100477 to XY). 
We thank Yuanqing Zhang and Lumin Yang for generously sharing their knowledge. 
We also thank the anonymous reviewers for their constructive suggestions on this manuscript.

{\small
\bibliographystyle{ieee_fullname}
\bibliography{egbib}
}

\newpage
\appendix
\section*{Appendix}

\section{Sampling from SG Sampling Function}
As described in Sec.~3.3, we utilize the Spherical Gaussian (SG) distribution sampling method to improve the Monte Carlo ray sampling efficiency: %
\begin{equation}
    \label{eq:result}
    p_{SG}(\boldsymbol{w}_i) = 
    \sum_{k=1}^M 
    a_k 
    \frac{\lambda_k}{2\pi(1-e^{-2\lambda_k})}
    e^{\lambda_k(\boldsymbol{w}_i \cdot \boldsymbol{\xi}_k - 1)}
    ,
\end{equation}
where $\boldsymbol{\xi}, \lambda, \boldsymbol{\mu} \in \boldsymbol{\Theta}_E$ are SG parameters of environment illumination, \ie, lobe axis, lobe sharpness and lobe amplitude of SG respectively.

When sampling, we first utilize the probability $a_k$ to decide which Gaussian component to draw from, then draw $\boldsymbol{w}_i$ from the $k$-th Gaussian distribution.
In order to draw samples from the $k$-th Gaussian distribution, we apply inverse transform sampling \cite{pharr2016physically}, which employs uniform random variables and maps them to random variables of the target distribution.
We first transform the PDF of direction $\boldsymbol{w}_i$ to 1D marginal and conditional density functions of its spherical coordinate $\varphi$ and $\theta$.
Following \cite{pharr2016physically}, 
the joint PDF ${p}_k(\theta, \varphi)$ of spherical coordinate $\varphi$ and $\theta$ is derived as

\begin{equation}
{p}_k(\theta, \varphi) = c_k \sin\theta e^{\lambda_k(\cos\theta - 1)}.
\end{equation}
Hence, the marginal density function ${p}_k(\theta)$ of $\theta$ is
\begin{equation}
\begin{split}
    {p}_k(\theta) 
        &= \int_{0}^{2\pi} c_k \sin\theta e^{\lambda_k(\cos\theta - 1)} d\varphi \\
        &= 2\pi c_k \sin\theta e^{\lambda_k(\cos\theta - 1)}
        .
\end{split}
\end{equation}
As a result, the conditional density function ${p}_k(\varphi|\theta)$ of $\varphi$ is 
\begin{equation}
    {p}_k(\varphi|\theta) = \frac{{p}_k(\theta, \varphi)}{{p}_k(\theta)} =  \frac{1}{2\pi}.
\end{equation}

We then compute the cumulative distribution function (CDF) of the distribution, $P_k(\theta)$ and $P_k(\varphi|\theta)$:
\begin{equation}
\begin{split}
    P_k(\theta) 
        &= \int_{0}^{\theta} 2\pi c_k \sin t e^{\lambda_k(\cos t - 1)} dt \\
        &= \frac{2 \pi c_k}{\lambda_k}(1 - e^{\lambda_k(\cos\theta - 1)}),
\end{split}
\end{equation}
\begin{equation}
    P_k(\varphi|\theta) = \int_{0}^{\varphi}\frac{1}{2\pi} = \frac{\varphi}{2\pi}.
\end{equation}

According to inverse transform sampling, random variable $X = F^{-1}_{X}(u)$ has distribution $F_{X}(x)$, where $u$ is a random value generated from the standard uniform distribution.
Hence, to apply inverse transform sampling and draw a sample $\theta$ based on a uniformly distributed random number $u_1$, we solve for $P_k(\theta) = u_1$:
\begin{equation}
\begin{split}
    \frac{2 \pi c_k}{\lambda_k}(1 - e^{\lambda_k(\cos\theta - 1)}) = u_1 \\
\Rightarrow    \theta = \arccos( 1 + 
                \frac{1}{\lambda_k}
                \ln(1 - 
                    \frac{\lambda_k u_1}{2\pi c_k}
                ) )
    .
\end{split}
\end{equation}
In a similar way, we can draw a sample $\varphi$ based on a uniformly distributed random value $u_2$ as:
\begin{equation}
    \varphi = 2 \pi u_2.
\end{equation}

\section{Comparison of albedo results of real scenes.}
\begin{figure}
\centering
\includegraphics[width=\columnwidth]{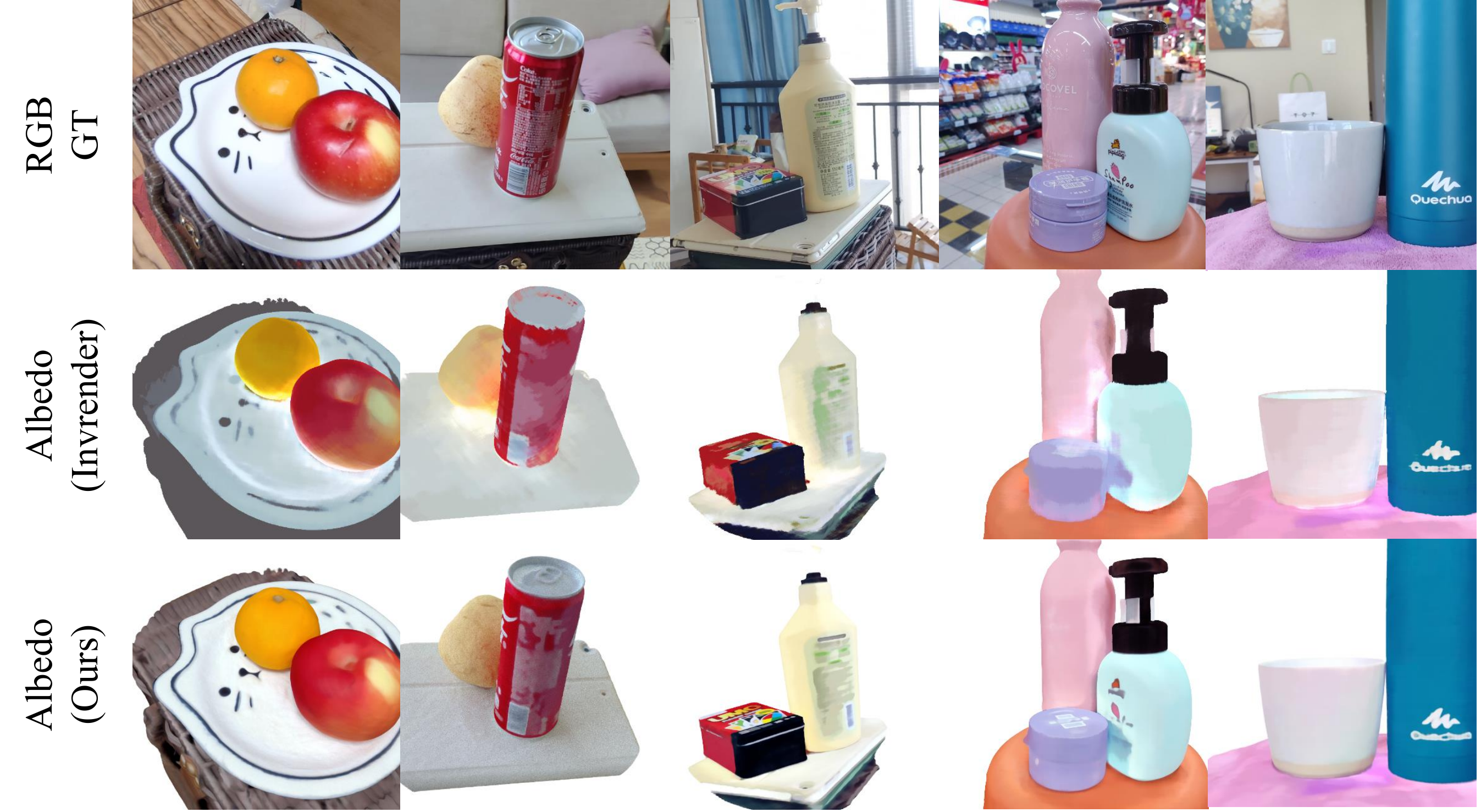}\\
\caption{
    Comparison of albedo results of real scenes.
}
\label{fig:real} 
\end{figure}

\cref{fig:real} illustrates the comparison between our method and Invrender in recovering diffuse materials in real scenes.
Our method outperforms Invrender in modeling indirect illumination, which helps to avoid baking indirect illumination into the albedo and causing incorrect brightness of certain  areas.

\begin{figure*}
\centering
\includegraphics[width=\linewidth]{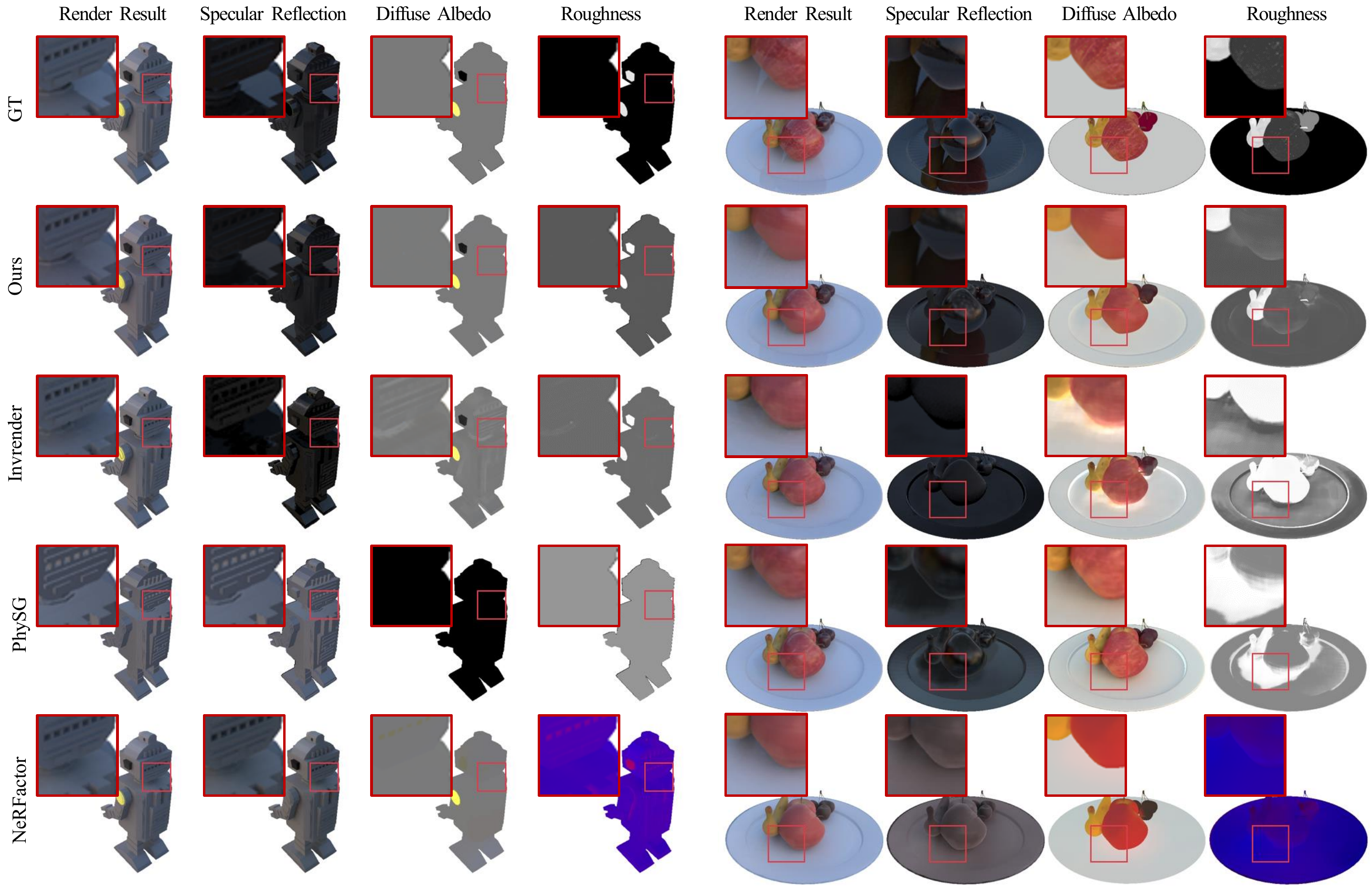}\\
\caption{
    Additional results of synthetic scenes.
}
\label{fig:synthetic} 
\end{figure*}

\begin{figure*}[tbp]
\centering
\includegraphics[width=1.0\textwidth]{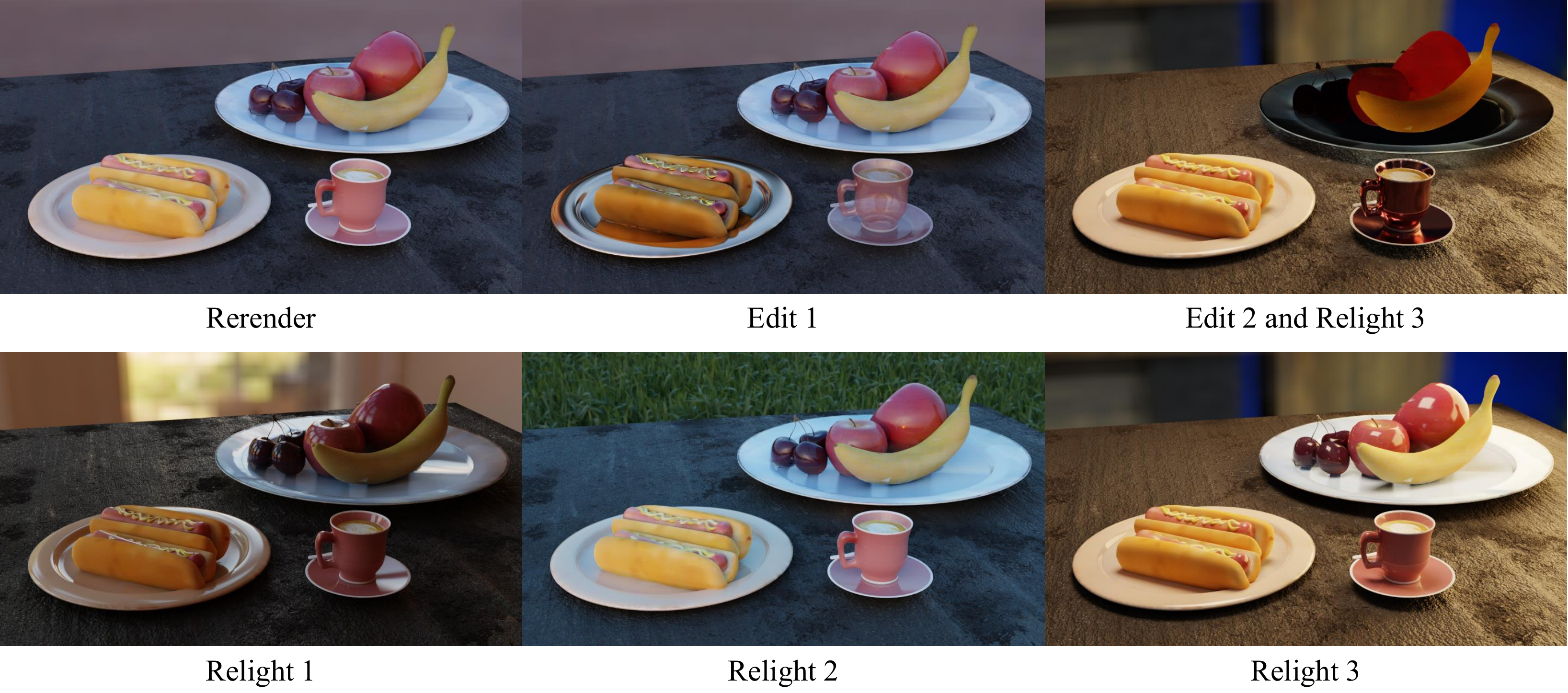}
\caption{
    \textbf{Results of Scene Manipulation in Blender.}
    We present material editing and relighting results in Blender~\cite{blender2018} of our recovered models, \ie, hotdog, coffee, and fruits.
    The results are rendered by Blender Cycles at 2048 ssp.
}
\label{fig:exp-blender} 
\end{figure*}

\section{Additional Results of synthetic scenes.}
\cref{fig:synthetic} shows qualitative comparisons results of the other synthetic scenes.

\section{Scene Manipulation in Blender}
Our method supports further scene manipulation in graphics engines. 
We convert our recovered material properties to image textures and import them into Blender~\cite{blender2018} based on OpenMVS~\cite{openmvs2020}.
In \cref{fig:exp-blender}, we present the material editing and relighting results of our recovered models, hotdog, coffee, and fruits.
Note that there are some biases of image textures caused by OpenMVS.

\end{document}

%% file: tab-baseline_comparison.tex
\begin{tabular}{|c|c|ccc|ccc|ccc|}
    \hline
    \multirow{2}{*}{Method}                              & Roughness        & \multicolumn{3}{c|}{Aligned Diffuse Albedo}             & \multicolumn{3}{c|}{View Synthsis Specular RGB}         & \multicolumn{3}{c|}{View Synthsis RGB}                  \\ \cline{2-11} 
                                                     & MSE $\downarrow$ & PSNR $\uparrow$  & SSIM $\uparrow$ & LPIPS $\downarrow$ & PSNR $\uparrow$  & SSIM $\uparrow$ & LPIPS $\downarrow$ & PSNR $\uparrow$  & SSIM $\uparrow$ & LPIPS $\downarrow$ \\ \hline
    NeRFactor \cite{zhang2021nerfactor} & -                & 21.8857          & 0.9159          & 0.0953             & 19.2751          & 0.8695          & 0.1147             & 29.9826          & 0.9597          & 0.0475             \\
PhySG \cite{zhang2021physg}         & 0.0481           & 19.7933          & 0.8988          & 0.1109             & 26.7784          & 0.9025          & 0.0693             & 31.0425          & \textbf{0.9642} & \textbf{0.0436}    \\
Invrender \cite{zhang2022modeling}  & 0.0464           & 27.4026          & 0.9426          & 0.0914             & 26.1370          & 0.9035          & 0.0831             & 30.8743          & 0.9616          & 0.0490             \\
\textbf{Ours}                       & \textbf{0.0065}  & \textbf{28.1094} & \textbf{0.9516} & \textbf{0.0845}    & \textbf{34.2930} & \textbf{0.9608} & \textbf{0.0416}    & \textbf{31.0909} & 0.9586          & 0.0528             \\ \hline

\end{tabular}